\pdfoutput=1

\documentclass[11pt]{article}

\usepackage[final]{acl}

\usepackage{times}
\usepackage{latexsym}

\usepackage[T1]{fontenc}

\usepackage[utf8]{inputenc}

\usepackage{microtype}

\usepackage{inconsolata}

\usepackage{bbm}
\usepackage{graphicx}
\usepackage{multirow, booktabs}
\usepackage{amsmath}
\usepackage{subfigure}
\usepackage{verbatim}
\usepackage{fancyvrb}
\usepackage{hyperref}
\newbox\verbbox
\title{Investigating Multi-Hop Factual Shortcuts in Knowledge Editing of \\Large Language Models}

\author{Tianjie Ju\textsuperscript{1}, 
Yijin Chen\textsuperscript{1}, 
Xinwei Yuan\textsuperscript{2}, 
Zhuosheng Zhang$^*$\textsuperscript{1}, 
Wei Du\textsuperscript{1}, \\ 
{\bf Yubin Zheng\textsuperscript{1}, 
Gongshen Liu\thanks{Corresponding authors.}\textsuperscript{1}} \\
\textsuperscript{1}School of Electronic Information and Electrical Engineering, Shanghai Jiao Tong University \\
\textsuperscript{2}School of Cyberspace Security, Southeast University\\
\texttt{\{jometeorie, st.czzz\}@sjtu.edu.cn, symor@seu.edu.cn,}\\
\texttt{\{zhangzs, dddddw, zybhk21, lgshen\}@sjtu.edu.cn}}

\begin{document}
\maketitle
\begin{abstract}

Recent work has showcased the powerful capability of large language models (LLMs) in recalling knowledge and reasoning. However, the reliability of LLMs in combining these two capabilities into reasoning through multi-hop facts has not been widely explored. This paper systematically investigates the possibilities for LLMs to utilize shortcuts based on direct connections between the initial and terminal entities of multi-hop knowledge. We first explore the existence of factual shortcuts through Knowledge Neurons, revealing that: (i) the strength of factual shortcuts is highly correlated with the frequency of co-occurrence of initial and terminal entities in the pre-training corpora; (ii) few-shot prompting leverage more shortcuts in answering multi-hop questions compared to chain-of-thought prompting. Then, we analyze the risks posed by factual shortcuts from the perspective of multi-hop knowledge editing. Analysis shows that approximately 20\% of the failures are attributed to shortcuts, and the initial and terminal entities in these failure instances usually have higher co-occurrences in the pre-training corpus. Finally, we propose erasing shortcut neurons to mitigate the associated risks and find that this approach significantly reduces failures in multiple-hop knowledge editing caused by shortcuts. Code is publicly available at \href{https://github.com/Jometeorie/MultiHopShortcuts}{https://github.com/Jometeorie/MultiHopShortcuts}.


\end{abstract}

\section{Introduction}

Large Language Models (LLMs) such as ChatGPT \citep{ChatGPT} and LLaMA-2 \citep{LLaMA-2}, have impressive world knowledge modeling and reasoning capabilities within their parameters \citep{LLM_survey,hao-etal-2023-reasoning}. 
When leveraging these two capabilities, it is intuitively anticipated that LLMs should be capable of reliably answering multi-hop knowledge questions without any difficulty \citep{multi-hop_QA}. 

Nonetheless, the underlying reasoning processes of LLMs in responding to multi-hop knowledge questions have not received thorough investigation. Ideally, an LLM would systematically derive each single-hop answer and culminate in the correct result. However, in reality, LLMs may leverage factual shortcuts learned from pre-training corpora to directly obtain the final answer without performing intermediate reasoning.

For conventional multi-hop question answering, the consistency of the final endpoints of shortcuts and multi-hop reasoning results may not cause risks and could even remain unnoticed. However, with the constant evolution of world knowledge, knowledge editing techniques are garnering increased attention \citep{survey_for_ke}. 
After knowledge editing, factual shortcuts in multi-hop scenarios may cause significant inconsistency.

\begin{figure}
  \centering
  \includegraphics[width=0.48\textwidth]{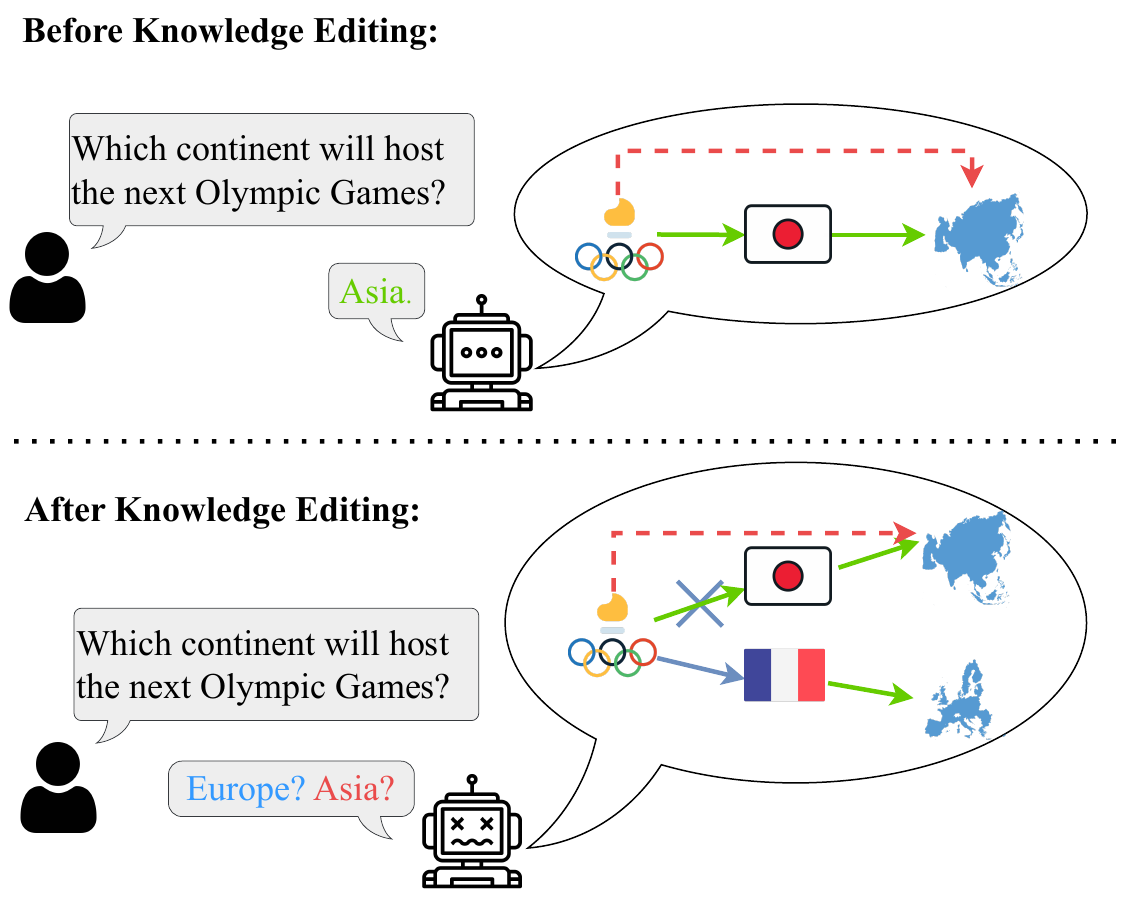}
  \caption{An illustrative example of a multi-hop factual shortcut in LLMs. The LLM may have directly encoded multi-hop knowledge (red) during the pre-training phase, which results in inconsistencies after a single-hop knowledge editing.}
  \label{fig:intro}
\end{figure}

Figure~\ref{fig:intro} illustrates the potential pitfalls associated with factual shortcuts. 
During the pre-training phase, an LLM may have forged a direct association between the next Olympic Games and Asia. 
Consequently, when queried with the prompt: \textit{``Which continent will host the next Olympic Games''}, the LLM might bypass the need for reasoning about the country and can directly furnish the correct answer. 
However, applying knowledge editing to the LLM, e.g., updating the host country of the Olympic Games to France, can expose a vulnerability.
The persistence of the established shortcut may lead the LLM to consistently output ``\textit{Asia}'' as the host continent even after the change, instead of the correct ``\textit{Europe}'', thereby impeding the success of multi-hop knowledge editing.

In this paper, we systematically investigate the possibilities for LLMs to utilize factual shortcuts based on direct connections between the initial and terminal entities of multi-hop knowledge. Firstly, we rethink and formalize the process through which LLMs reason about multi-hop knowledge. \textbf{We introduce the hypothesis that LLMs may leverage factual shortcuts from pre-training corpora to facilitate cross-step reasoning}. 

Then, we deeply explore the existence of factual shortcuts. We conduct a frequency analysis of co-occurrences between the initial subject and terminal object of multi-hop knowledge instances in pre-training corpora. Additionally, we employ Knowledge Neurons \citep{KN} to quantify the overlap between the activated neurons for multi-hop questions and all single-hop questions. A low degree of overlap suggests that the reasoning pattern of LLMs in response to multi-hop questions is inconsistent with that of single-hop questions, indicating the presence of shortcuts. Our experiments on multi-hop knowledge reveal that: 

(i) Few-shot questions exhibit more shortcuts in comparison to chain-of-thought questions, suggesting that \textbf{LLMs often arrive at multi-hop knowledge answers using unexpected cross-step reasoning patterns}. 

(ii) Knowledge instances with a higher co-occurrence frequency between initial subjects and terminal objects tend to have more shortcuts, indicating \textbf{a strong correlation between the existence of multi-hop factual shortcuts and the word frequencies learned by LLMs during pre-training phase}.

Additionally, to provide insights into the potential risks associated with multi-hop factual shortcuts, we conduct a detailed analysis of the reasons behind the failures in multi-hop knowledge editing. We find that \textbf{approximately 20\% of the failure instances are attributed to multi-hop factual shortcuts}. Furthermore, \textbf{shortcut failure instances often exhibit higher co-occurrence frequencies of the initial and terminal entities}, providing compelling evidence that the presence of shortcuts may disrupt the multi-hop reasoning consistency of LLMs after knowledge editing.

Finally, we explore the feasibility of employing Knowledge Neurons to eliminate factual shortcuts. We erase crucial neurons associated with factual shortcuts that co-occurred more than 10 times in the pre-training corpus. Results show that \textbf{the failure rate of multiple-hop knowledge editing caused by shortcuts significantly decreased, leading to an overall improvement in the success rate after our erasing approach}. We hope this work can facilitate increased interest in exploring the multi-hop reasoning capabilities of LLMs and constrain reasoning shortcuts during the pre-training stage.

\section{Rethinking the Multi-Hop Knowledge}
\label{sec: Rethinking the Multi-Hop Knowledge}

A basic fact can be formulated as a single-hop knowledge tuple $t = \left(s, r, o \right)$ with a subject ($s$), a relation ($r$), and an object ($o$). For each query, we ask the LLM if the object is correct given the subject and the relation $\mathbbm{1} \left\{f \left(T \left(s, r \right) \right) = o \right\}$, where $f$ and $T$ denote the outputs of the LLM and the prompt template for splicing $s$ and $o$ into a cloze-style form.

In this paper, we mainly focus on the multi-hop knowledge, which comprises a chain of single-hop knowledge:
\begin{equation}
    \mathcal{T} = \left< \left( s_1, r_1, o_1 \right), ..., \left( s_n, r_n, o_n \right)\right>, 
\end{equation}
where $s_i = o_{i-1}$. For each query, we directly ask the LLM if the terminal object is correct given the initial subject and the chain relation $\mathbbm{1} \left\{f \left(T \left(s_1, r_{mul} \right) \right) = o_n \right\}$, where $r_{mul} = r_1 \rightarrow ... \rightarrow r_n$. This question can also be formulated as asking the LLM of the knowledge tuple $t_{mul} = (s_1, r_{mul}, o_n)$, which proves unproblematic in general multi-hop question-answering, as $t_{mul}$ and $\mathcal{T}$ share the same endpoint $o_n$.

However, $t_{mul}$ is in fact a shortcut, treating a chain of relations as a separate composite relation. If a knowledge-editing approach is employed to modify the intermediate entity $o_i$ to $o_i^*$, the final answer of $\mathcal{T}$ will be altered. Since $t_{mul}$ overlooks the intermediate entity, its answer remains unaffected by knowledge editing.

It is also reasonable from the cause-and-effect perspective. For a two-hop knowledge $e_1 \rightarrow e_2 \rightarrow e_3$, it requires first deducing the intermediate entity $e_2$ to obtain the correct output $e_3$. Any other reasoning path, such as $e_1 \rightarrow e_3$ and $e_1 \rightarrow e_4 \rightarrow e_3$, does not conform to the causal relationship.

Taking the multi-hop question of ``\textit{Which continent will host the next Olympic Games}'' as an illustrative example, if we edit the knowledge of the ``\textit{country}'' from Japan to France, according to the chain-relation reasoning, the ``\textit{continent}'' hosting the Olympic Games should be converted to Europe. However, if a composite relation is employed, the ``\textit{continent}'' would remain unchanged despite alterations in the ``\textit{country}''.

A causal LLM probably encodes such composite knowledge during the pre-training phase. The initial subject $s_1$ and the terminal object $o_n$ are likely to have direct associations in the corpus. Still taking the example above, an LLM may have learned the knowledge \textit{(the next Olympic Games, continent of the country, Asia)} from the corpus directly, neglecting the causal relationship between the country and the continent to which it belongs. Therefore, for multi-hop knowledge, LLMs may potentially arrive at the correct answer through step-wise reasoning, but it is more likely that they memorize the outdated answer by leveraging the co-occurrence relationships in the pre-training corpus.

\section{Exploring the Existence of Factual Shortcuts}
\label{sec: Exploring the Existence of Factual Shortcuts}

In this section, we explore the extent of shortcuts in multi-hop question-answering. Concretely, we first validate the correlation between multi-hop shortcuts and the word frequency in the pre-training corpus. Then, we locate crucial neurons in single-hop, few-shot, and chain-of-thought question-answering tasks to further elucidate the degree of potential factual shortcuts.

\subsection{Probing Shortcuts in Pre-training Corpus}

Our analysis centers specifically on the \textsc{mquake-cf-3k} dataset released by \citet{MQUAKE}. It comprises 1,000 two-hop, 1,000 three-hop, and 1,000 four-hop instances of multi-hop question-answering for knowledge editing extracted from Wikidata \citep{wikidata}. Essential information for one sample from the dataset is shown in Appendix~\ref{Dataset}. we compute the crucial neurons of the first question within the 'questions' key alongside the entirety of questions within the 'single\_hops' key.
It can also be adopted in subsequent sections for further investigating potential risks introduced by these multi-hop factual shortcuts.

Considering that the existence of factual shortcuts may drive from pre-training corpora, we first compute the frequency of co-occurrence of the initial subject $s_1$ and the terminal object $o_n$ among these 3,000 items of knowledge on Wikipedia (20231101-en) and Dolma corpus(v1\_6-sample) respectively. The Wikipedia dataset contains approximately 6.41M rows of text, while the Dolma dataset contains roughly 10 billion tokens. We chose these two corpora due to their comprehensive coverage of global knowledge and their frequent utilization as a significant component in the pre-training corpora for most LLMs. If $s_1$ and $o_n$ co-occur within the same paragraph, it is highly plausible that the LLM establishes a direct connection between them during the pre-training phase.

\begin{figure}
  \centering
  \includegraphics[width=0.48\textwidth]{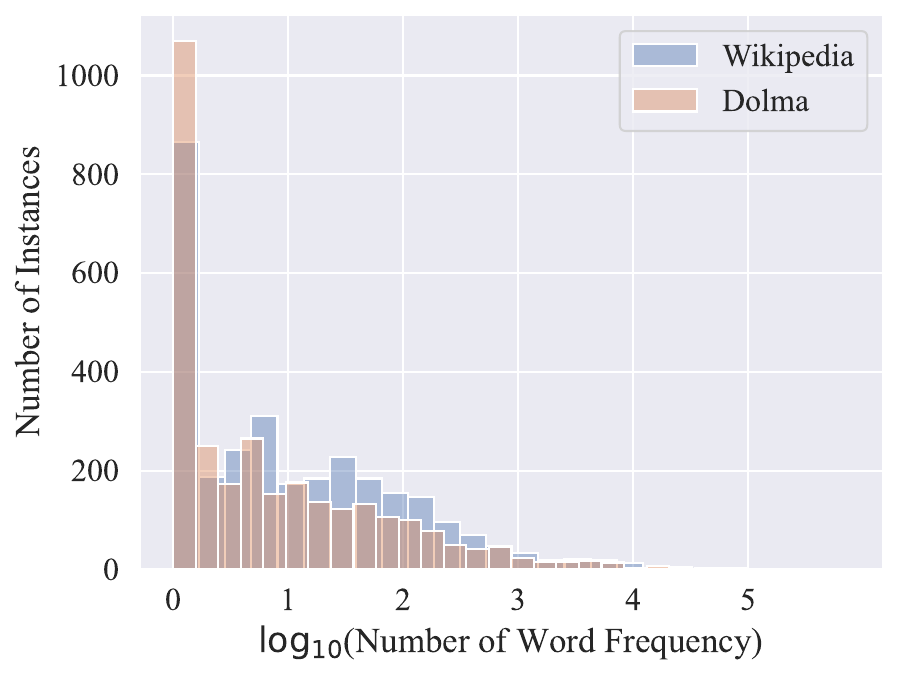}
  \caption{Frequency analysis of multi-hop knowledge shortcuts in Wikipedia and Dolma.}
  \label{fig:count_res}
\end{figure}

We first conduct a frequency analysis of the occurrences of these multi-hop knowledge shortcuts in the Wikipedia corpus (Figure~\ref{fig:count_res}). It can be observed that more than 2/3 of instances exhibited various degrees of shortcuts, with some even appearing over 10,000 times. This indicates that \textbf{certain pieces of knowledge exhibit significant multi-hop shortcuts}, which could potentially influence the reasoning processes of LLMs.

We've also conducted the same frequency analysis on the Dolma corpus (Figure~\ref{fig:count_res}). It can be observed that the co-occurrence rate distribution of vocabulary in the Dolma dataset is \textbf{similar} to that in the Wikipedia dataset. We calculate the Pearson Coefficient of the co-occurrence rate between the two datasets, and the result is 0.74. The similarity in the co-occurrence rates between these two datasets indicates the strong correlation between the two corpora, and therefore, Wikipedia can be chosen as an approximate corpus.

Moreover, we select several examples with high and low frequencies for illustration (Table~\ref{tab: Examples of multi-hop knowledge}) in the Wikipedia corpus. It can be observed that instances with high frequency exhibit a clear, direct connection between $s_1$ and $o_n$. For instance, ``\textit{Twitter}'' is inherently strongly associated with ``\textit{the United States}'', obviating the need to think about the country of citizenship of ``\textit{Twitter's CEO}''. In contrast, there is no apparent connection between ``\textit{Jerry Rivers}'' and ``\textit{Donald Trump}'', necessitating the prior derivation of the nationality of ``\textit{Jerry Rivers}'' to arrive at the correct answer. Since ``\textit{Jerry Rivers}'' and ``\textit{Donald Trump}'' rarely co-occur in the pre-training corpus, LLMs may not contain factual shortcuts related to such multi-hop knowledge. 

\begin{table*}
 \centering
 \resizebox{\linewidth}{!}{
 \begin{tabular}{ccccc}\toprule
    Subject ($s_1$) & Object ($o_n$) & Multi-Hop Question & $f_{\textrm{Wikipedia}}$ &  $f_{\textrm{Dolma}}$ \\\midrule
    Rhode Island & English & Which languages are spoken, written, or signed in Rhode Island as the head of government there? & 42754 & 21279 \\
    Twitter & United States of America & What is the country of citizenship of Twitter's CEO? & 35435 & 205862 \\
    Fanta & Atlanta & What is the location of the headquarters of the manufacturer of Fanta? & 25834 & 72910 \\
    \midrule
    Jerry Rivers & Donald Trump & Who is the head of state of the country whose citizen is Jerry Rivers? & 0 & 0 \\
    Alvar Aalto & Mikael Agricola & Who is the creator of the content in the language or languages spoken by Alvar Aalto? & 0 & 0\\
    Nick Rimando & London & What is the capital of the country where the sport of Nick Rimando's position is originated? & 0 & 0\\
    \bottomrule
 \end{tabular}}
 \caption{Examples of multi-hop knowledge with high and low frequency of co-occurrence of $s_1$ and $o_n$, where $f$ denotes the frequency of co-occurrence in the pre-trained corpus.}
 \label{tab: Examples of multi-hop knowledge}
\end{table*}

\subsection{Quantifying Shortcuts Using Knowledge Neurons}

\begin{figure}
  \centering
  \includegraphics[width=0.48\textwidth]{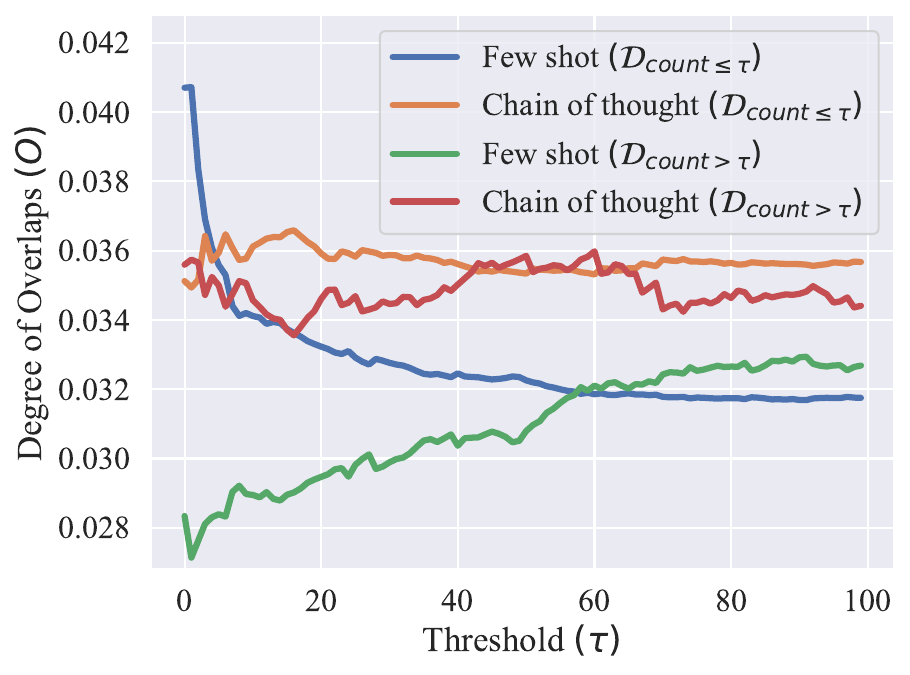}
  \caption{The degree of overlaps employed by GPT-J in handling multi-hop questions and all single-hop questions with varying word frequencies in pre-training corpora under few-shot prompts and chain-of-thought prompts. It is expected that the instances from $\mathcal D_{\textrm{count} > \tau}$ contain more potential factual shortcuts.}
  \label{fig: neurons}
\end{figure}

\paragraph{Methods.} The presence of multi-hop factual shortcuts may result in a divergence in the reasoning mechanisms employed by the LLM when responding to multi-hop questions as opposed to directly answering individual single-hop questions. To quantify the disparities, we employ Knowledge Neurons (KN) proposed by \citet{KN} to locate crucial neurons activated by the LLMs when responding to various questions. Specifically, it gradually changes each neuron $w_i^{(l)}$ stored in FFN from 0 to its original value $\overline{w}_i^{(l)}$ and meanwhile integrates the gradients. We use the Riemann approximation as a substitution for continuous integrals:
\begin{equation}
    \tilde{\textrm{Attr}} (w_i^{(l)}) = \frac{\overline{w}_i^{(l)}}{m} \sum_{k=1}^m \frac{\partial P(\frac{k}{m} \overline{w}_i^{(l)})} {\partial w_i^{(l)}},
\end{equation}
where $P(w_i^{(l)}) = p(y|x, w_i^{(l)} = \hat w_i^{(l)})$ is the probability of the correct answer predicted by the LLM when changing the value of neuron $w_i^{(l)}$ to $\hat w_i^{(l)}$, and $m$ is the number of the approximation steps. We choose neurons with attribution values larger than $v$ as crucial neurons reflecting LLM decision-making patterns:
\begin{equation}
    \mathbf N = \left\{ w_i^{(l)} | \textrm{Attr}(w_i^{(l)}) > v \right\}.
\end{equation}

\begin{figure}
  \centering
  \includegraphics[width=0.48\textwidth]{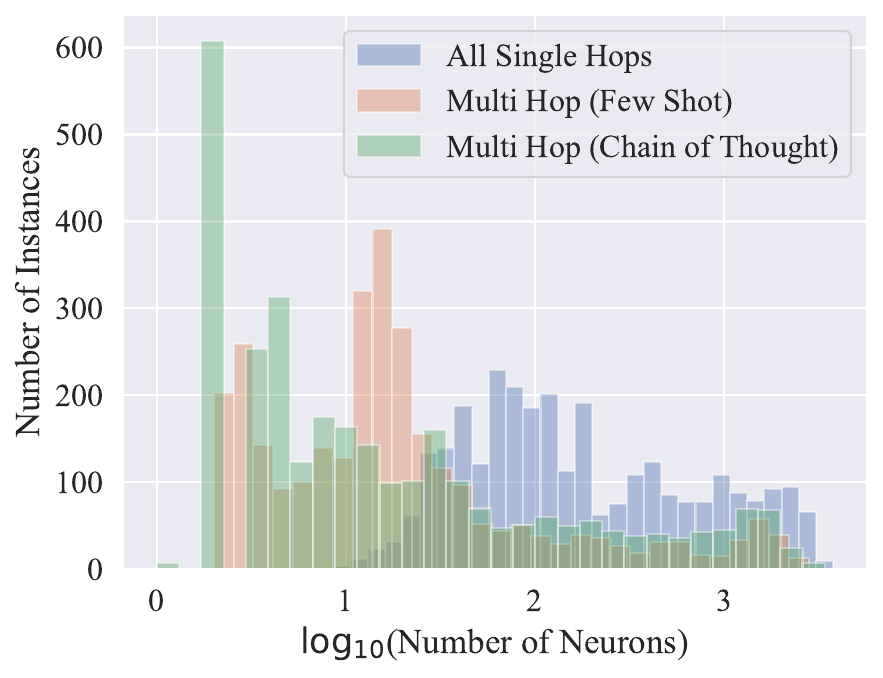}
  \caption{Distribution of the number of activated neurons in GPT-J across different questions.}
  \label{fig: neurons_freq}
\end{figure}

In this paper, we set $m$ to 20 and the attribution threshold $v$ to 0.2. In the scenario of a multi-hop question devoid of any shortcuts, it should ideally encompass a broader array of crucial neurons inherent to single-hop questions, except those specifically dedicated to lower-level components such as lexical and syntactic neurons. Hence, we define $O$ as the degree of overlap between the reasoning patterns of multi-hop knowledge answers and all single-hop knowledge answers:
\begin{equation}
    O = \frac{|\mathbf N_{\mathcal{T}} \cap \mathbf N_{t_{mul}}|}{|\mathbf N_{\mathcal{T}}|},
\end{equation}
where $\mathbf N_{\mathcal{T}}$ denotes the intersection of crucial neurons for all single-hop questions, $N_{t_{mul}}$ denotes the set of crucial neurons for multi-hop questions. A higher degree of overlap indicates that LLM's reasoning patterns for answering multi-hop questions are more closely aligned with those for answering single-hop questions.

It is noteworthy to emphasize that our objective does not entail the precise localization of neurons storing knowledge; rather, we aim to discern the decision-making processes of the LLMs across various questions.
Despite \citet{KN_rebuttal}'s skepticism regarding whether neurons uncovered by KN in the FFN truly constitute ``knowledge'', these neurons may store intricate ``token expression patterns'' that can still elucidate the LLM's decision-making processes.

We separately evaluate the degree of shortcuts in few-shot and chain-of-thought multi-hop questions. 
All single-hop questions and few-shot multi-hop questions utilize the same demonstrations, while chain-of-thought multi-hop questions employ prompts with similar semantics. 

For all single-hop questions, we adopt the few-shot prompt shown in Table~\ref{tab: few-shot prompt}. Subsequently, we locate crucial neurons based on the probability of correct answer output by the LLM following the "A:" prefix.

For multi-hop questions, we adopt both the few-shot and chain-of-thought prompts. The few-shot prompt mirrors that of single-hop questions, while the chain-of-thought prompt is constructed with semantically approximate expressions. We require the LLM to articulate its reasoning process upon receiving the question. Then we locate crucial neurons based on the probability of correct answer output by the LLM following the "Answer:" prefix (see Table~\ref{tab: chain-of-thought prompt}). Both prompts are provided in Appendix~\ref{sec: Prompts for Knowledge Neurons}). 

Besides, we partition the original dataset $\mathcal D_o$ into two subsets $\mathcal D_{\textrm{count} \leq \tau}$ and $\mathcal D_{\textrm{count} > \tau}$ based on word frequencies, where $\tau$ represents the threshold for word frequencies. 
We compute the degree of shortcuts for GPT-J (Figure~\ref{fig: neurons}).

\begin{table*}
 \centering
 \resizebox{\linewidth}{!}{
 \begin{tabular}{lcccccccccccccc}\toprule
               & & \multicolumn{4}{c}{$\mathcal{S}$} & $\mathcal{F_{\textrm{single}}}$ & \multicolumn{4}{c}{\textbf{$\mathcal{F_{\textrm{shortcut}}}$}} & \multicolumn{4}{c}{$\mathcal{F_{\textrm{other}}}$} \\
               \cmidrule(lr){3-6}\cmidrule(lr){7-7}\cmidrule(lr){8-11}\cmidrule(lr){12-15}
               & & $i = 1$ & $i = 2$ & $i = 3$ & Sum & Sum & \textbf{i = 1} & \textbf{i = 2} & \textbf{i = 3} & \textbf{Sum} & $i = 1$ & $i = 2$ & $i = 3$ & Sum \\
               \midrule
    \multirow{3}{*}{GPT-J (6B)} & MEND & 4.27 & 4.53 & 14.17 & 22.97 & 33.03 & \textbf{3.93} & \textbf{3.17} & \textbf{11.87} & \textbf{18.97} & 5.40 & 4.97 & 33.47 & 43.84 \\
     & ROME & 2.07 & 2.30 & 4.57 & 8.94 & 39.87 & \textbf{3.13} & \textbf{2.27} & \textbf{9.17} & \textbf{14.57} & 3.17 & 3.63 & 41.13 & 47.93 \\
    & MEMIT & 2.17 & 1.97 & 4.87 & 9.01 & 33.37 & \textbf{4.10} & \textbf{3.63} & \textbf{11.47} & \textbf{19.20} & 4.20 & 4.00 & 43.27 & 51.47 \\
    \midrule
    \multirow{3}{*}{LLaMA-2 (7B)} & MEND & 7.40 & 4.80 & 7.77 & 19.97 & 43.57 & \textbf{5.63} & \textbf{5.70} & \textbf{9.63} & \textbf{20.96} & 5.90 & 6.20 & 26.90 & 39.00 \\
     & ROME & 5.33 & 3.00 & 3.83 & 12.16 & 25.37 & \textbf{6.30} & \textbf{6.17} & \textbf{11.67} & \textbf{24.14} & 6.80 & 7.00 & 44.67 & 58.47 \\
    & MEMIT & 5.13 & 3.60 & 3.83 & 12.56 & 32.00 & \textbf{6.00} & \textbf{5.47} & \textbf{10.17} & \textbf{21.64} & 6.20 & 7.13 & 40.33 & 53.66 \\
    \bottomrule
 \end{tabular}}
 \caption{The percentage of successful $(\mathcal{S})$ and failed $(\mathcal{F})$ multi-hop knowledge edits, where $i$ denotes the frequency of success or failure within the three queries, "Sum" denotes the cases with at least one success or failure. We mainly focus on failures caused by factual shortcuts $(\mathcal{F_{\textrm{shortcut}}})$.}
 \label{tab: Times of successes and failures}
\end{table*}

\paragraph{Main Results.} It can be observed that \textbf{the LLM adheres to a greater extent to reasoning patterns overlapping with those for single-hop questions under the chain-of-thought prompt}. This observation suggests that the chain-of-thought prompt indeed serves to induce the LLM to engage in step-wise reasoning. It also aligns with our hypothesis that LLMs tend to prioritize the utilization of latent multi-hop factual shortcuts, relinquishing them only when explicitly prompted to engage in step-wise reasoning. Furthermore, the instances within $\mathcal D_{\textrm{count} > \tau}$ exhibit lower degrees of reasoning overlap, suggesting that \textbf{LLMs indeed learn the shortcut associations between $s_1$ and $o_n$, with word frequencies significantly influencing the strength of these shortcuts}.

Interestingly, although the overlap rates vary across different scenarios, their values remain low. We analyze the distribution of the number of activated knowledge neurons for different instances (Figure~\ref{fig: neurons_freq}). Since single-hop knowledge typically involves 2-4 questions, the number of activated neurons is an order of magnitude higher than that for multi-hop questions. Moreover, activated neurons, in addition to reflecting the inherent knowledge, may also be influenced by factors such as the lexical and syntactic aspects of sentences. Hence, the reasoning overlap rates tend to be maintained at a low value.

For the hyper-parameter $v$, We randomly select 500 instances at $v=0.1$ and $v=0.3$ for experiments (Table~\ref{tab:ablation}). The results show the same trend in the figure with $\tau$ as the X-axis. Although the magnitude of the values varies, this is because as $v$ decreases, more neurons will be considered crucial neurons, which will significantly increase the size of the denominator $|\mathbf N_{\mathcal{T}}|$, leading to a decrease in the overlap rate. Nevertheless, our conclusions are correct under different parameters: (i) the strength of factual shortcuts is highly correlated with the frequency of co-occurrence of initial and terminal entities in the pre-training corpora; (ii) few-shot prompting leverage more shortcuts in answering multi-hop questions compared to chain-of-thought prompting.

\begin{table}[h]
    \centering
    \resizebox{\columnwidth}{!}{
        \begin{tabular}{lccc}
            \hline
            & $v = 0.1$ & $v = 0.2$ & $v = 0.3$ \\
            \hline
            $\mathcal{D}_{\textrm{count} \leq 10}$ (Few Shot) & 3.09 & 3.41 & 7.67 \\
            $\mathcal{D}_{\textrm{count} > 10}$ (Few Shot) & 2.40 & 2.89 & 6.76 \\
            $\mathcal{D}_{\textrm{count} \leq 10}$ (Chain of Thought) & 3.01 & 3.61 & 8.08 \\
            $\mathcal{D}_{\textrm{count} > 10}$ (Chain of Thought) & 2.73 & 3.46 & 7.82 \\
            \hline
        \end{tabular}
    }
    \caption{Ablation studies on the hyper-parameter $v$}
    \label{tab:ablation}
\end{table}

\section{Exploring the Potential Risks of Factual Shortcuts}
\label{sec: Exploring the Potential Risks of Factual Shortcuts}

\begin{figure*}[htb]
    \centering
	\subfigure[GPT-J (6B)]{
	\begin{minipage}{0.48\linewidth}
    	\label{GPT-J_failure}
    	\centering
    	\includegraphics[width=1\linewidth]{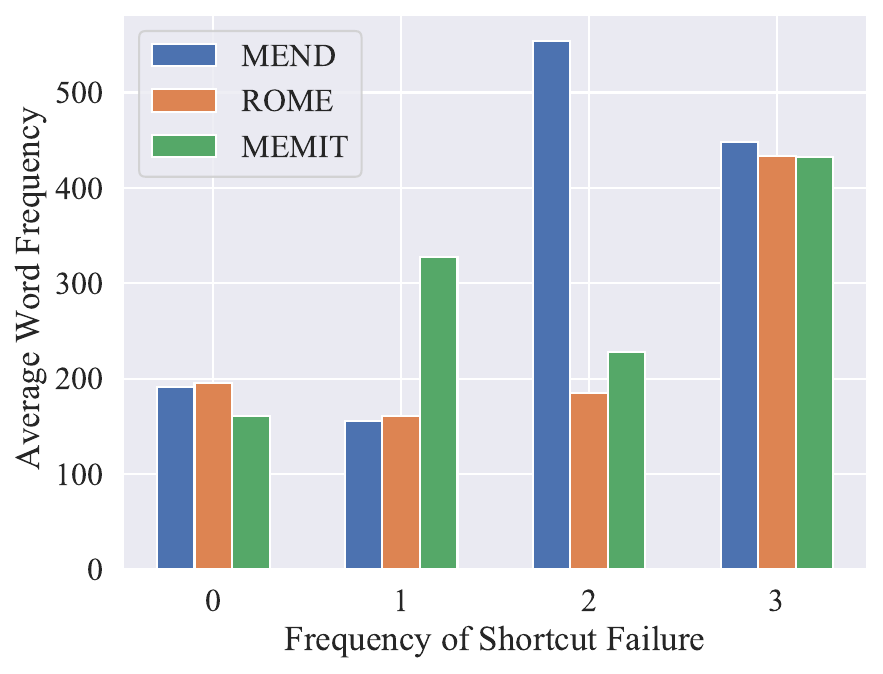}
	\end{minipage}
	}
        \subfigure[LLaMA-2 (7B)]{
	\begin{minipage}{0.48\linewidth}
    	\label{LLaMA_failure}
    	\centering
    	\includegraphics[width=1\linewidth]{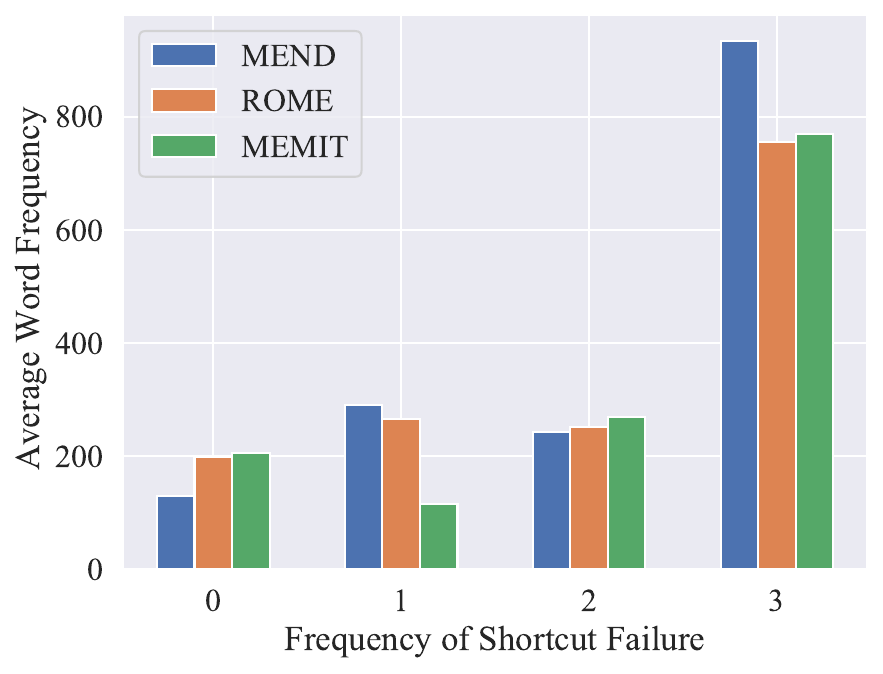}
	\end{minipage}
	}
    \caption{The average co-occurrence frequency of $s_1$ and $o_n$ in the pre-training corpus. The horizontal axis represents the number of occurrences of shortcut failures across three queries.}
    \label{fig: failure}
\end{figure*}

While these shortcuts may not have a significant impact on the results in general multi-hop question answering, their potential risks can be magnified in the context of knowledge editing. \citet{MQUAKE} have observed poor performance of LLMs in multi-hop knowledge editing. In this section, we will specifically analyze the reasons for the failure of multi-hop knowledge editing, particularly under the influence of shortcuts.

Concretely, we employ various knowledge editing methods to modify single-hop knowledge instances in \textsc{mquake-cf-3k} and pose three different multi-hop questions about the edited knowledge. Subsequently, we quantify the effects of various knowledge editing methods and categorize error instances into three categories.

\paragraph{Failure Categories.} We consider three key categories of failures. The first category of failure stems from the unsuccessful editing of single-hop knowledge. We designate the set of failures in this category as $\mathcal{F_{\textrm{single}}}$. The second and third categories are built upon the assumption of successfully editing all single-hop knowledge instances, yet the LLM still fails to answer multi-hop questions correctly. The second category signifies cases where the answer to multi-hop knowledge questions remains the same as the original unedited answer. We denote this set as $\mathcal{F_{\textrm{shortcut}}}$. Given that all single-hop questions can be answered correctly, the persistence of the original result in multi-hop questions indicates the existence of shortcuts. The third category involves the LLM providing alternative incorrect answers, potentially arising from hallucinations or other reasons. We denote this set as $\mathcal{F_{\textrm{other}}}$.

For each multi-hop edited knowledge, we interrogate the LLM with three distinct multi-hop questions. All multi-hop questions are prefixed with the same few-shot template comprising 16 demonstrations, which is consistent with the setup of \citet{MQUAKE}. We calculate the percentage of editing successes $(\mathcal{S})$ and failures $(\mathcal{F})$ within three questions. Detailed experimental settings can be seen in the Appendix~\ref{sec: Experimental Details}.

\paragraph{Main Results.} Table~\ref{tab: Times of successes and failures} presents the analysis results.
Consistent with the findings of \citet{MQUAKE}, knowledge editing algorithms exhibit catastrophic failures when addressing multi-hop factual questions, with only approximately 10\%-20\% of instances avoiding complete errors across three queries. $\mathcal{F_{\textrm{single}}}$ stems from the editing failure of LLMs in addressing single-hop questions. Since multi-hop questions may necessitate more than one edit, it may be slightly higher than the edit-wise failure rate. $\mathcal{F_{\textrm{other}}}$ may originate from the insufficient reasoning capabilities of LLMs or the hallucinations generated during editing. While we utilize few-shot prompts instead of chain-of-thought prompts to expose factual shortcuts, it may also increase $\mathcal{F_{\textrm{other}}}$.

It is noteworthy that $\mathcal{F_{\textrm{shortcut}}}$ also constitutes a significant proportion. This type of failure implies that LLMs respond with old ground truth for multi-hop questions while capable of correctly answering single-hop questions after all edits. In other words, shortcuts enable LLMs to conveniently utilize the $r_{mul}$ hard-coded during the pre-training phase to directly obtain results, without genuinely engaging in multi-hop knowledge reasoning. \textbf{Experiments indicate that these factual shortcuts are prevalent across various knowledge types in LLMs}.

To further investigate the connection between shortcut failures and falsely learned relations in the pre-training corpus, we analyze the relationship between the average co-occurrence frequency of entities and the occurrence frequency of shortcut failures (Figure~\ref{fig: failure}). We observe that \textbf{instances with higher occurrences of shortcut failures, particularly those with three failures, exhibit higher word co-occurrence frequencies between $s_1$ and $o_n$}. This suggests that LLMs are highly likely to leverage the multi-hop knowledge hardcoded during the pre-training phase as reasoning shortcuts. The presence of these factual shortcuts significantly diminishes the reliability and plausibility of LLMs' reasoning. In the context of multi-hop knowledge editing, the LLMs are easily entangled in the confusion between old shortcut knowledge and new multi-hop knowledge.

\section{Reducing Multi-Hop Factual Shortcuts}
\label{sec: Reducing Multi-Hop Factual Shortcuts}

The existence of multi-hop factual shortcuts reveals the unreliability of current LLMs' reasoning and increases the risk of failures in multi-hop knowledge editing. Since these shortcuts represent knowledge hardcoded into LLMs during the pre-training phase, it is challenging to eliminate these factual shortcuts fundamentally unless there are substantial changes in the pre-training phase.

\begin{figure}
  \centering
  \includegraphics[width=0.48\textwidth]{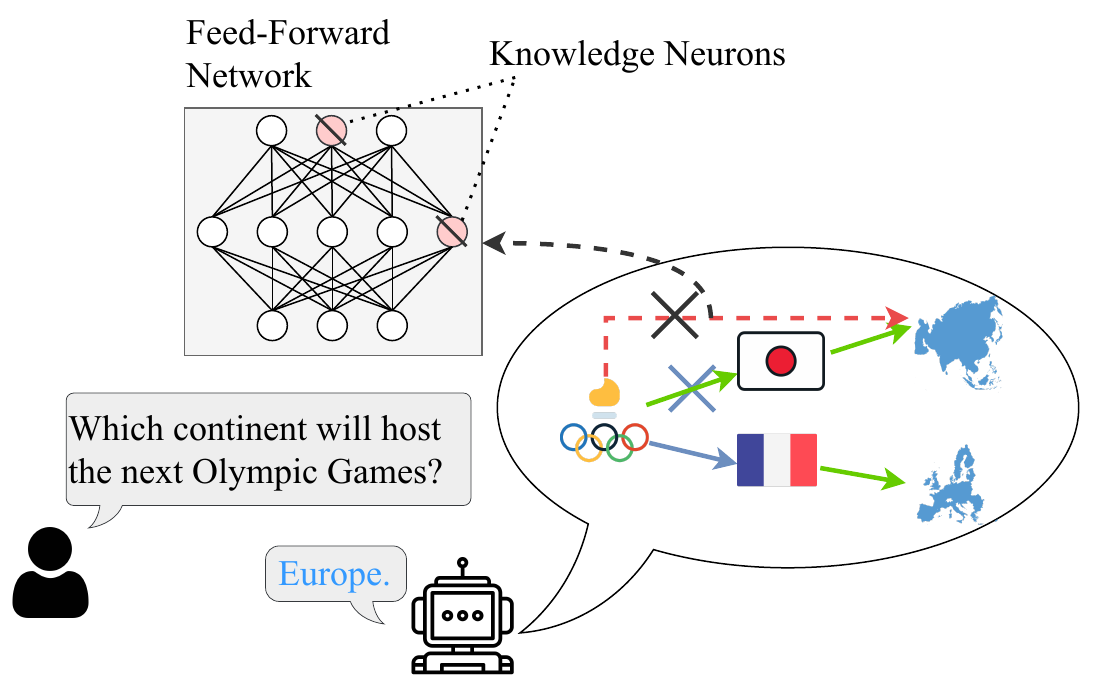}
  \caption{An illustrative example for reducing multi-hop factual shortcuts.}
  \label{fig: erase}
\end{figure}

\paragraph{Methods.} To reduce the risks of multi-hop factual shortcuts and further validate the hypotheses presented in this paper, we adopt a simple yet effective method inspired by \citet{KN} to erase these shortcuts (Figure~\ref{fig: erase}). Compared to Figure~\ref{fig:intro}, we erase crucial neurons related to the red factual shortcuts, compelling the LLM to answer the continent that will host the next Olympic Games using the correct path of reasoning after knowledge editing.

Specifically, we use the integral gradient algorithm to locate the crucial neurons associated with multi-hop knowledge questions and set them to zero. For each piece of multi-hop knowledge, we query with three questions to obtain the intersection of crucial neurons. Based on the previous experiments (Figure~\ref{fig: neurons}), we posit that multi-hop knowledge with a co-occurrence frequency exceeding 10 exhibits evident shortcuts. Consequently, we proceeded to eliminate these multi-hop factual shortcuts from the dataset $\mathcal D_{\textrm{count} > 10}$. We compute the percentage of editing success $(\mathcal{S})$ and shortcut failure rate $(\mathcal{F}_{\textrm{shortcut}})$ for multi-hop knowledge editing before and after the erase of factual shortcuts, respectively.

\begin{table*}
 \centering
 \small
  \setlength{\tabcolsep}{7.2pt}
 \begin{tabular}{lcccccccccc}\toprule
               & & & \multicolumn{4}{c}{$\mathcal{S} \uparrow$} & \multicolumn{4}{c}{\textbf{$\mathcal{F_{\textrm{shortcut}}} \downarrow$}} \\
               \cmidrule(lr){4-7}\cmidrule(lr){8-11}
               & & & $i = 1$ & $i = 2$ & $i = 3$ & Sum & $i = 1$ & $i = 2$ & $i = 3$ & Sum \\
               \midrule
    \multirow{6}{*}{GPT-J (6B)} & \multirow{2}{*}{MEND} & Before Erasing & 4.46 & 5.13 & \textbf{19.56} & 29.15 & \textbf{4.08} & \textbf{4.18} & 17.57 & 25.83 \\
     &  & After Erasing & \textbf{5.79} & \textbf{5.41} & 18.42 & \textbf{29.62} & 4.75 & 4.84 & \textbf{15.76} & \textbf{25.35} \\
     \cmidrule(lr){2-11}
     & \multirow{2}{*}{ROME} & Before Erasing & 2.09 & \textbf{2.94} & \textbf{8.64} & 13.67 & 2.75 & \textbf{3.23} & 12.82 & 18.80 \\
     &  & After Erasing & \textbf{4.47} & \textbf{2.94} & 8.36 & \textbf{15.77} & \textbf{2.57} & 3.33 & \textbf{11.62} & \textbf{17.52} \\
     \cmidrule(lr){2-11}
     & \multirow{2}{*}{MEMIT} & Before Erasing & 1.61 & \textbf{2.94} & 7.98 & 12.53 & 4.27 & 5.60 & 16.05 & 25.92 \\
     &  & After Erasing & \textbf{3.32} & 2.66 & \textbf{8.07} & \textbf{14.05} & \textbf{3.23} & \textbf{4.65} & \textbf{14.25} & \textbf{22.13} \\
    \midrule
    \multirow{6}{*}{LLaMA-2 (7B)} & \multirow{2}{*}{MEND} & Before Erasing & \textbf{9.21} & 5.79 & \textbf{9.97} & \textbf{24.97} & \textbf{6.27} & \textbf{7.03} & 17.76 & 31.06 \\
     &  & After Erasing & 8.36 & \textbf{6.08} & 9.31 & 23.75 & 7.79 & 8.17 & \textbf{5.51} & \textbf{21.47} \\
     \cmidrule(lr){2-11}
     & \multirow{2}{*}{ROME} & Before Erasing & 5.98 & 4.65 & \textbf{7.03} & 17.66 & \textbf{6.84} & 6.93 & 18.33 & 32.10  \\
     &  & After Erasing & \textbf{7.50} & \textbf{4.75} & 6.93 & \textbf{19.18} & 7.03 & \textbf{6.36} & \textbf{11.97} & \textbf{25.36}  \\
     \cmidrule(lr){2-11}
     & \multirow{2}{*}{MEMIT} & Before Erasing & 5.60 & 4.84 & \textbf{7.12} & 17.46 & \textbf{6.08} & 6.17 & 17.09 & 29.34 \\
     &  & After Erasing & \textbf{8.36} & \textbf{5.03} & 6.74 & \textbf{20.13} & 6.36 & \textbf{5.88} & \textbf{9.88} & \textbf{22.12} \\
    \bottomrule
 \end{tabular}
 \caption{Success rate and shortcut failure rate of multi-hop knowledge editing before and after the erase of factual shortcuts on $\mathcal D_{\textrm{count} > 10}$.}
 \label{tab: erase}
\end{table*}

\paragraph{Main Results.} Table~\ref{tab: erase} presents the success rate and shortcut failure rate of multi-hop knowledge editing before and after the erase of factual shortcuts on $\mathcal D_{\textrm{count} > 10}$. 
Compared to Tabel~\ref{tab: Times of successes and failures}, both the success rate and shortcut failure rate of multiple-hop knowledge editing have increased on $\mathcal D_{\textrm{count} > 10}$. 
The result implies that instances with factual shortcuts are inherently more amenable to editing, yet the presence of factual shortcuts also entails a higher level of risk for these instances. Thus, these latent factual shortcuts are far more harmful than we realize.

Furthermore, \textbf{the erasing of shortcuts can significantly reduce the risks associated with shortcut failures, leading to an appreciable improvement in the success rate of multi-hop knowledge editing}. Due to the incapacity of knowledge editing methods to address shortcut knowledge $t_{mul}$, inconsistencies arise in LLMs' reasoning results. By erasing neurons corresponding to $t_{mul}$, we ensure that LLMs reason along the correct path, thereby enhancing the success rate.

However, despite the efficacy of this approach in mitigating the risk posed by factual shortcuts to specific knowledge, it cannot serve as a comprehensive solution to the problem. Due to the ubiquitous nature of such shortcuts, it is impractical to review and erase crucial neurons for every multi-hop knowledge. Fundamentally, to attain a trustworthy LLM with genuine multi-hop reasoning capabilities, it is imperative to address the issue at the pre-training stage to explore improved pre-training methodologies.

\section{Related Work}
In this section, we discuss two lines of research that are key to our work: knowledge editing and multi-hop reasoning. 

\subsection{Knowledge Editing} 
Numerous studies have explored efficient knowledge editing methods for LLMs, seeking resolutions to challenges arising from outdated knowledge. One prevalent and intuitive approach involves employing external memorization, wherein new knowledge is incorporated through external context or parameters, without necessitating modifications to the LLM weights \citep{SERAC, CALINET, T-Patcher, IKE, MQUAKE}. While these approaches are simple and effective in ensuring consistency, the substantial influx of supplementary knowledge may result in redundancy and low timeliness at a later stage \citep{survey_for_ke}.

Another line of work focuses on directly updating the LLM parameters. Some investigations are dedicated to constrained fine-tuning \citep{RecAdam, PPA} or meta-learning \citep{PPA, MEND}, which update the full parameters of LLMs. The other investigations involve a preliminary stage of knowledge localization before editing, premised on the assumption that knowledge is stored in the form of key-value memories within the two-layer Feedforward Neural Network (FFN) \citep{memory_key_value}. \citet{KN} located and refined knowledge neurons (KN) through integral gradients \citep{IG}. \citet{ROME} et al. proposed the Rank-One Model method (ROME) to insert new knowledge in a specific FFN layer, while MEMIT \citep{MEMIT} further extended address scenarios of mass editing.

While the effectiveness of single-hop knowledge editing has been thoroughly investigated, there is a notable dearth of attention given to multi-hop knowledge editing. \citet{MQUAKE} systematically focused on this issue by introducing the multi-hop knowledge editing evaluation benchmarks \textsc{mquake-cf} and \textsc{mquake-t}. Their findings revealed catastrophic performance degradation of existing knowledge editing methods. In this paper, we further investigate and elucidate the repercussions stemming from the presence of reasoning shortcuts in multi-hop knowledge editing.

\subsection{Multi-Hop Reasoning} Multi-hop reasoning is often seen as a weakness for LLMs \citep{reasoning_servey}. Early efforts commonly employed in-context prompting, which involves the provision of few input-output demonstrations to LLMs \citep{few_shot_1, few_shot_2, few_shot_3}. This approach enables LLMs to solve problems through reasoning implicitly. However, its effectiveness diminishes significantly when confronted with multi-hop questions \citep{few_shot_fail}. To incentivize LLMs to engage in explicit multi-hop reasoning, the concept of \textit{chain-of-thought} was introduced by \citet{chain_of_thought}. It encourages the LLM to think step by step and output intermediate deductive steps \citep{chain_of_thought_survey}. In this paper, we elucidate the process by which LLMs handle multi-hop question-answering from the perspective of factual shortcuts. We provide evidence that the chain-of-thought prompting compels LLMs to attend to the single-hop knowledge more faithfully.

\section{Conclusion}

In this paper, we systematically explore the latent factual shortcuts that LLMs may employ when answering multi-hop knowledge questions. We first demonstrate the strong correlation between the strength of factual shortcuts and the co-occurrence of the initial subject and the terminal object in pre-training corpora. Then, we delve into the potential risks introduced by these shortcuts in the context of multi-hop knowledge editing. Our exploration reveals that approximately 20\% of failures can be attributed to factual shortcuts, particularly in instances characterized by high co-occurrences within pre-training corpora. Finally, we propose a straightforward yet efficient approach to mitigate shortcut failures in multi-hop knowledge editing by selectively erasing shortcut neurons. We advocate for increased research efforts directed towards exploring the true boundaries of LLMs in the realm of multi-hop reasoning, emphasizing the need to better constrain shortcut generation during the pre-training phase.

\section*{Acknowledgements}
This work is partially supported by the National Key R\&D Program of China under No. 2023YF3303800 and the Joint Funds of the National Natural Science Foundation of China under No.U21B2020.

\section*{Limitations}

We posit that Wikipedia serves as a comprehensive repository of global knowledge, thus making it a suitable substitute for the entirety of the pre-training corpora. However, despite our exhaustive traversal of the Wikipedia dataset to calculate the co-occurrence frequencies of initial and terminal entities, it is noteworthy that the pre-training corpora for LLMs often extend beyond the confines of this dataset. This potential discrepancy may introduce inaccuracies in statistical outcomes. We advocate for future investigations to extend statistical analyses to more expansive corpora.


For the erasing of factual shortcuts, our primary objective is to further substantiate the potential risks associated with these shortcuts, and the observed improvement in editing success rates after erasing serves to support this assertion. However, it is imperative to recognize that this approach functions as a mitigative measure, as the complete eradication of factual shortcuts through post-hoc removal is unattainable. A genuine and thorough elimination of factual shortcuts must be initiated during the pre-training phase, involving the alignment of LLMs' multi-hop reasoning capabilities with human-level proficiency.

Finally, due to space and resource constraints, we only conduct detailed experiments on GPT-J (6B) and LLaMA-2 (7B) and do not encompass all publicly accessible LLMs, such as PaLM~\citep{palm}, OPT~\citep{opt}, and Pythia~\citep{pythia}. 
We encourage future research to undertake comprehensive experiments on a broader spectrum of LLMs. 

\section*{Ethical Statement}

We conduct a reassessment of the multi-hop reasoning capabilities of LLMs and demonstrate that the presence of factual shortcuts may compromise the consistency of results in multi-hop knowledge editing. Since the approach itself is unbiased and all experiments are conducted on publicly available datasets, we believe that our work creates no potential ethical risk. Additionally, all use of existing artifacts is consistent with their intended use in this paper.

However, we have exposed the indiscriminate use of shortcuts by LLMs during multi-hop reasoning, raising concerns regarding their genuine reasoning capabilities. LLMs struggle to engage in step-wise reasoning akin to human cognitive processes, and the potential for parameter confusion may arise following the assimilation of new knowledge. These factors contribute to our perplexity concerning the black-box nature of LLMs and apprehensions regarding their application in security-sensitive domains. We advocate for more rigorous ethical scrutiny and improvements in LLMs to ensure alignment with the human reasoning process.


\bibliography{custom}

\clearpage
\appendix

\section{Dataset}
\label{Dataset}

We select the \textsc{mquake-cf-3k} dataset as the primary focus for exploration in this paper. It comprises 3,000 multi-hop English knowledge questions extracted from Wikipedia along with a corresponding knowledge editing task. We present essential information for one sample from the dataset (Table~\ref{tab: dataset}). For Section~\ref{sec: Exploring the Existence of Factual Shortcuts}, we compute the crucial neurons of the first question within the 'questions' key, alongside the entirety of questions within the 'single\_hops' key. For Section~\ref{sec: Exploring the Potential Risks of Factual Shortcuts}, we adopt knowledge editing methods of all knowledge encapsulated within the 'requested\_rewrite' key. Furthermore, we augment the original dataset by introducing a new key, labeled as 'shortcut\_frequency', which denotes the frequency of co-occurrence in the pre-training corpus between the initial subject and the terminal object for each instance.

\setbox\verbbox=\vbox{
\begin{Verbatim}[commandchars=\\\{\}]
case_id: 16
requested_rewrite: [
    \{
        prompt: \{\} is a citizen of
        target_new: Latvia,
        target_true: United States of America,
        subject: Jack Dorsey,
        question: What is the country of citizenship of Jack Dorsey?
    \}
]
questions: [
    What is the country of citizenship of Twitter's CEO?
    From which country does Twitter's CEO hold citizenship?
    Which country's citizenship is held by the CEO of Twitter?
]
answer: United States of America
answer_alias: ...
new_answer: Latvia
new_answer_alias: ...
\textbf{shortcut_frequency: 35435}
single_hops: [
    \{
        question: Who is the chief executive officer of Twitter?
        cloze: The chief executive officer of Twitter is
        answer: Jack Dorsey
        answer_alias: ...
    \}
    \{
        question: What is the country of citizenship of Jack Dorsey?
        cloze: Jack Dorsey is a citizen of
        answer: United States of America
        answer_alias: ...
    \}
]
new_single_hops: [
    \{
        question: Who is the chief executive officer of Twitter?
        cloze: The chief executive officer of Twitter is
        answer: Jack Dorsey
        answer_alias: ...
    \}
    \{
        question: What is the country of citizenship of Jack Dorsey?
        cloze: Jack Dorsey is a citizen of
        answer: Latvia
        answer_alias: ...
    \}
]
\end{Verbatim}
}

\begin{table*}
\centering
\begin{tabular*}{\hsize}{l}
\hline
\box\verbbox \\
\hline
\end{tabular*}
\caption{Critical information for a sample in the multi-hop knowledge editing dataset \textsc{mquake-cf-3k}. We have added the 'shortcut\_frequency' key to the original dataset to store the frequency of shortcuts appearing in Wikipedia.}
\label{tab: dataset}
\end{table*}

\section{Prompts for Knowledge Neurons}
\label{sec: Prompts for Knowledge Neurons}

We employ prompt templates similar to that utilized by \citet{MQUAKE} for finding crucial neurons. Given the substantial computational overhead associated with Knowledge Neurons, we adopt a 2-shot prompt, which is already sufficient for the LLM to comprehend the task and furnish accurate responses. The few-shot prompt and chain-of-thought prompt are shown in Table~\ref{tab: few-shot prompt} and Table~\ref{tab: chain-of-thought prompt}.

\begin{table*}
    \centering
        \begin{tabular*}{\hsize}{l}
            \hline
            Q: Who is the spouse of the US president? A: Jill Biden\\
            Q: In which country is the company that created Nissan 200SX located? A: Japan\\
            Q: [Input Question] A: [Output Answer]\\
            \hline
        \end{tabular*}
    \caption{The few-shot prompt for Knowledge Neurons.}
    \label{tab: few-shot prompt}
\end{table*}

\begin{table*}
    \centering
        \begin{tabular*}{\hsize}{l}
            \hline
            Question: Who is the spouse of the US president?\\
            Thoughts: The US president is Joe Biden. The spouse of Joe Biden is Jill Biden.\\
            Answer: Jill Biden.\\\\

            Question: In which country is the company that created Nissan 200SX located?\\
            Thoughts: Nissan 200SX was created by Nissan. Nissan is located in the country of Japan.\\
            Answer: Japan.\\\\

            Question: [Input Question]\\
            Thoughts: [Output Thoughts]\\
            Answer: [Output Answer]\\
            \hline
        \end{tabular*}
    \caption{The chain-of-thought prompt for Knowledge Neurons.}
    \label{tab: chain-of-thought prompt}
\end{table*}

\section{Experimental Details}
\label{sec: Experimental Details}

\subsection{Language Models}
\label{sec: Language Models}

Our experiments are conducted on GPT-J (6B) \citep{GPT-J} and LLaMA-2 (7B) \citep{LLaMA-2}. The selection of GPT-J is motivated by the alignment with the pre-existing work on knowledge editing \citep{ROME, MEMIT, MQUAKE}, while opting for LLaMA-2 is motivated by its status as a recent, prominent open-source LLM representative, providing a robust reflection of the current capabilities of LLMs. We use the huggingface package \citep{HuggingFace} for the specific implementation.

\subsection{Knowledge Editing}
\label{sec: Knowledge Editing}

We use the cloze-style statement templates for knowledge editing, which is consistent with the previous studies. We employ the EasyEdit package \citep{EasyEdit} for the specific implementation. All licenses of these packages allow us for normal research use. The detailed specifics of the three knowledge editing methods that are employed in our training are as follows.

\paragraph{MEND.} MEND \citep{MEND} trains a lightweight model editor network to produce edits to the LLM's weight when provided with the standard fine-tuning gradient. We train our editor network on the ZsRE dataset \citep{zsre} with a maximum number of training steps of 100,000. We set the learning rate scale to be 1.0 during inference. All experiments edit the MLP weights in the last 3 Transformer blocks.

\paragraph{ROME.} ROME \citep{ROME} stands out as a popular method for knowledge localization and editing. It introduces a based on corruption and restoration to identify relevant layers storing knowledge. Subsequently, it inserts new knowledge by key selection and value optimization in the corresponding feed-forward network (FFN) layer. We perform the intervention at layer 5 for GPT-J (6B) and 6 for LLaMA-2 (7B). We compute the second-order momentum statistics using 100,000 examples of Wikitext in \texttt{fp32}. For the remaining hyperparameters, we adopt the default values specified in \citet{ROME}.

\paragraph{MEMIT.} MEMIT \citep{MEMIT} is a subsequent work to ROME, designed to handle extensive knowledge edits. In this paper, we perform the intervention at layer \{3, 4, 5, 6\} for GPT-J (6B) and \{4, 5, 6, 7\} for LLaMA-2 (7B). We also compute the covariance statistics using 100,000 examples of Wikitext in \texttt{fp32}. For the remaining hyperparameters, we adopt the default values specified in \citet{MEMIT}.

\subsection{Computational Budget}

For all the experiments mentioned in this paper, we use one Nvidia A100-SXM4 GPU with 80GB memory. We spend about 100, 200, and 250 GPU hours exploring the existence of factual shortcuts (Section~\ref{sec: Exploring the Existence of Factual Shortcuts}), exploring the potential risks of factual shortcuts (Section~\ref{sec: Exploring the Potential Risks of Factual Shortcuts}) and reducing multi-hop factual shortcuts (Section~\ref{sec: Reducing Multi-Hop Factual Shortcuts}).

\end{document}